\PassOptionsToPackage{backref=page}{hyperref}

\documentclass{hld2026}

\usepackage[capitalize,noabbrev]{cleveref}
\usepackage{enumitem}

\usepackage{wrapfig}

\renewcommand*{\backref}[1]{}
\renewcommand*{\backrefalt}[4]{%
    \ifcase #1\or Cited on page~#2.\else Cited on pages~#2.\fi%
}

\title{Structure and Scale in Simplicial Sequence Modelling}

\hldauthor{%
    \Name{Matthew Farrugia-Roberts}
    \Email{matthew@far.in.net}\\
    \addr{Department of Computer Science, University of Oxford}
}

\begin{document}

\maketitle

\begin{abstract}%
    Modern large-scale deep learning exhibits two striking empirical phenomena:
    behavioural scaling laws (predictable performance gains with increasing
    scale) and emergent mechanisms (structured internal representations and
    circuits in deep neural networks).
    We hypothesise that these two phenomena are connected: that predictable
    changes in behaviour are the result of predictable changes in internal
    computational structure. In this paper, we report preliminary evidence of
    such a connection.
    We find a correlation between scaling patterns in performance and
    representations in small transformers trained to predict the outputs of a
    hidden Markov model, for which residual activations are known to linearly
    encode a belief distribution over latent states in a probability simplex.
\end{abstract}


\section{Introduction}
    
Through modern advancements in software
    \citep{Ciresan+2011,Krizhevsky+2012},
hardware
    \citep{Amodei+Hernandez2018},
and neural network architecture
    \citep{He+2016,Bahdanau+2015,Vaswani+2017},
it has become possible to train increasingly large neural networks on
increasingly large amounts of data
    \citep[e.g.,][]{Radford+2018,Radford+2019,Brown+2020,OpenAI2023,OpenAI2025}.
The resulting models have driven remarkable growth in applications of AI
throughout society
    \citep{Sajadieh+2026}.
While these applications have captured the world's attention, from the
perspective of a \emph{natural scientist,} the most interesting empirical
phenomena in recent AI history are the following:
\begin{enumerate}
    \item
        \textbf{Examples of interpretable internal structure:} Sometimes,
        trained neural networks store interpretable internal representations of
        relevant variables
            \citetext{e.g.,
                \citealp{Karpathy+2016},
                \citealp{Radford+2017},
                \citealp{Olah+2020},
                \citealp{Elhage+2022},
                \citealp{Li+2023},
                \citealp{Nanda+2023othello},
                \citealp{Shai+2024},
                \citealp{Riechers+2025},
                \citealp{Taufeeque+2024},
                \citealp{Bush+2025},
                \citealp{Engels+2025},
                \citealp{Gurnee+2026},
                \citealp{An+2026}%
            },
        and operate on these representations using interpretable circuits
            \citetext{e.g.,
                \citealp{Olah+2020},
                \citealp{Nanda+Lieberum2022},
                \citealp{Nanda+2023grokking},
                \citealp{Zhong+2023},
                \citealp{Elhage+2021},
                \citealp{Olsson+2022},
                \citealp{Wang+2023ioi},
                \citealp{Piotrowski+2025},
                \citealp{Taufeeque+2026}%
            }.
    
    \item
        \textbf{Principled variation in behaviour with scale:} The behaviour of
        trained neural networks follows predictable patterns, such as
        performance power laws, as a function of the scale of training (the
        number of data points, model parameters, or optimisation steps) over
        many orders of magnitude
            \citetext{e.g.,
                \citealp{Hestness+2017},
                \citealp{Rosenfeld+2020},
                \citealp{Kaplan+2020},
                \citealp{Hoffmann+2022},
                \citealp{Besiroglu+2024}%
            }.
\end{enumerate}
We hypothesise that these two phenomena are each reflections of a third
underlying phenomenon:
\begin{enumerate}[resume]
    \item
        \textbf{Principled variation in structure with scale:} the \emph{degree
        and kind of computational structure present inside} trained neural
        networks follow predictable patterns as a function of scale.
\end{enumerate}
Hypothetically, it is these changes in structure that drive changes in
behaviour---in contrast to models of scaling that view neural networks as
tabular function approximators
    \citep{Sharma+Kaplan2020,Hutter2021,Michaud+2023}.
Moreover, examples of interpretable mechanisms and representations appear when
the underlying changes in structure happen to surface as particularly
interpretable---a salient special case against a broader backdrop of
unexplained frontier model cognition
    \citep{Olah2024darkmatter,Nanda+2025}.

In this paper, we investigate this hypothesis by studying a known example of
structured representations from the mechanistic interpretability literature,
namely the simplicial sequence modelling setting of \citet{Shai+2024}.
In this setting, we show that with increasing scale (training steps), the
internal representations become increasingly refined, roughly following a power
law.

\section{Background}

\begin{figure}
    \centering
    \null\hfill
    \includegraphics[width=0.45\linewidth]{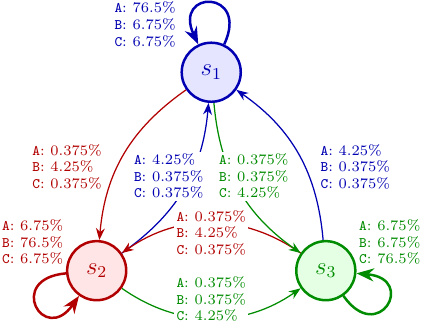}
    \hfill
    \includegraphics[trim={27 31 24 14},clip,width=0.41\linewidth]{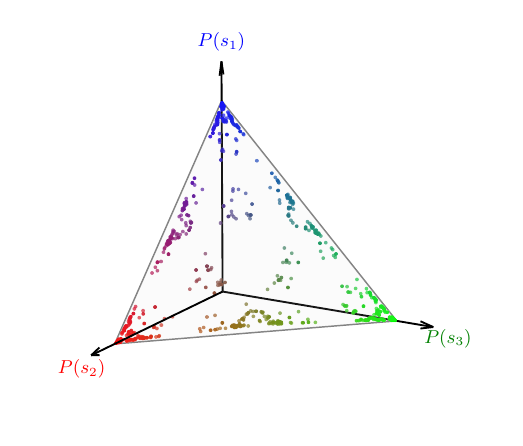}
    \hfill\null
    \caption{%
    \textbf{Simplicial sequence modelling.}
    \textit{Left:} Hidden Markov model sequence generator.
        Edge annotations denote probability of transitioning between latent
        states $\smash{s_1,s_2,s_3}$ while emitting symbols $\mathtt{A},
        \mathtt{B}, \mathtt{C}$.
    \textit{Right:} Probability simplex of belief distributions over latent
        states $\smash{s_1,s_2,s_3}$. Scattered points are Bayesian belief
        distributions from a sample of sequences.
    }
    \label{fig:mess3}
\end{figure}

We generated a data set of observation sequences of length 10 sampled from an
edge-emitting hidden Markov model with three latent states (\cref{fig:mess3},
left). The specific hidden Markov model is the ``mess3'' process studied by
\citet{Marzen+Crutchfield2017} with parameters $\alpha=0.85$, $x=0.05$.

Optimally predicting the next observation given 0--9 previous observations
requires computing the Bayesian posterior belief distribution over latent
states and averaging next-symbol probabilities over this uncertainty. These
belief distributions reside in the 2-simplex of distributions over three latent
states (\cref{fig:mess3}, right).

\citet{Shai+2024} trained transformers on this data distribution and found that
their transformers linearly encode Bayes-like belief distributions in final
layer activations. They found that these representations were less strongly
encoded at select mid-training checkpoints. We extend this work into an
in-depth mechanistic scaling experiment as follows.

\section{Experiments}
\label{sec:experiments}

We trained transformers to predict mess3 sequences using cross-entropy loss. We
used a fixed architecture with four residual transformer blocks, each with one
attention head of width~8 and one MLP layer of width~256 for a total of
approximately 142 thousand parameters.
We trained with stochastic mini-batch gradient descent using learning rate~0.01
and batches of 64~sequences.
We use a transformer training implementation in JAX \citep{Bradbury+2018}
derived from \citet{FarrugiaRoberts2026hijax}. Each training run took
approximately 3.7~hours per 10~million training steps on a TPU~v4-2 device.

\paragraph{Key variables.}

We scale the number of training steps while holding architecture, number of
parameters, and data generator constant. Compute scales linearly with training
steps, as does the number of unique training sequences (we generate fresh
sequences for every batch). We continually monitor the following observables
(for the first 1,000 steps, and thereafter at $\approx$100 steps per OOM).
\begin{enumerate}
    \item
        \textbf{Excess prediction loss:}
        Mean next-token prediction cross-entropy loss across a fixed held-out
        batch of 1024 sequences, minus Bayes-optimal cross-entropy on that
        batch.
    \item
        \textbf{Probe loss:}
        Mean squared error of a linear probe \citep{Alain+Bengio2018}
        reconstructing ground-truth Bayesian belief distributions from the
        transformer's final layer activations.
        (Mean taken across a fixed held-out batch of 1024 sequences,
        independent of sequences used to train the probe.)
\end{enumerate}

\begin{figure}[t!]
    \centering
    \includegraphics{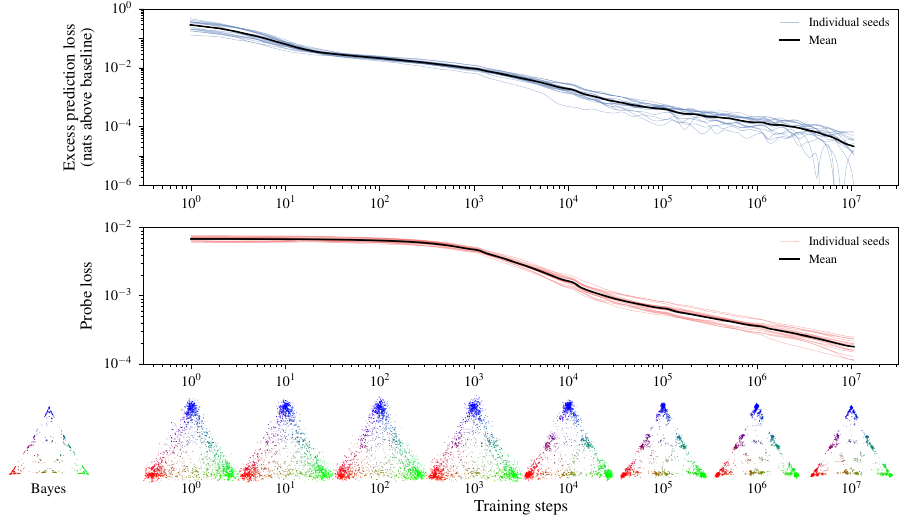}
    \caption{%
        \textbf{Performance/representation scaling of simplicial sequence
        models.}
        \textit{Top:} Per-step held-out test cross-entropy minus per-seed
        irreducible cross-entropy. Mean (black) of 16 seeds (blue), light
        log-Gaussian smoothing (see \cref{apx:smoothing}).
        \textit{Middle:} Mean squared error predicting Bayesian belief
        distributions from final-layer activations. Mean (black) of 16 seeds
        (red), light log-Gaussian smoothing.
        \textit{Bottom:} Bayesian posterior beliefs (``Bayes'') and
        reconstructed transformer beliefs at select training steps (1~seed).
    }
    \label{fig:results1}
\end{figure}

\paragraph{Representations improve smoothly with scale.}

We trained 16~transformers for 10~million training steps (10~times longer than
\citeauthor{Shai+2024}). \Cref{fig:results1} shows the resulting trends in
excess prediction loss and probe loss, along with a visualisation of belief
distributions reconstructed from activations via trained probes at select
steps.
We see that excess prediction loss continually improves with increasing compute
(roughly as a power law). Continuously measuring probe loss reveals that it
quickly also enters a regime of continual improvement (again, roughly as a
power law). The belief reconstruction visualisations show that the fractal
geometry is more crisply represented with scale.

\begin{figure}[t!]
    \centering
    \includegraphics[trim={0 1.5 2.5 3.5},clip]{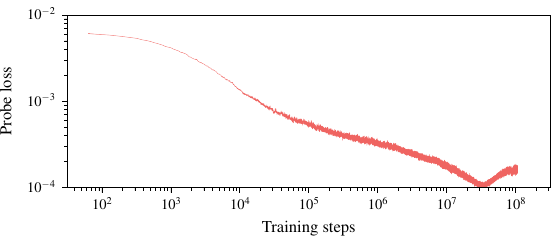}%
    \hfill%
    \includegraphics[trim={1 2.5 4 2.5}]{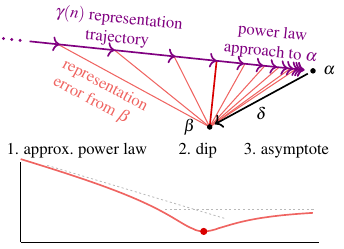}
    \caption{%
        \textbf{Long-run representation scaling in a simplicial sequence
        model.}
        \textit{Left:}
        Mean squared error predicting Bayesian belief distributions from
        final-layer activations (every 64~steps, no smoothing).
        \textit{Right:}
        Toy model of dynamics in representation subspace.
    }
    \label{fig:results2}
\end{figure}

\paragraph{Asymptotic versus ideal representations.}

We trained one transformer for 100~million steps (100~times longer than
\citeauthor{Shai+2024}). \Cref{fig:results2} shows that between 10~million and
100~million training steps, the probe loss is non-monotonic, turning around and
then apparently beginning to plateau.

If the transformer's representations were asymptotically approaching the ideal
Bayesian belief distributions, then we would expect monotonic scaling to
continue indefinitely.
However, failure to converge to the Bayesian ideal does not preclude
convergence to an alternative, realisable representation.
Indeed, finite-sized transformers cannot express the Bayesian
posterior exactly \citep{Piotrowski+2025}, so they might asymptote to a
realisable representation.
Below we provide one toy model of representation scaling that recovers the
power-law--dip--plateau pattern.

Let $\mathbb{R}^d$ be the joint space of final-layer transformer activations
within some residual subspace for each sequence in a large fixed batch. Let
    $\gamma : \mathbb{N} \to \mathbb{R}^d$
represent the trajectory of a transformer's activations in this subspace over
training, so that after
    $n \in \mathbb{N}$
training steps, the transformer's activations are
    $\gamma(n) \in \mathbb{R}^d$.
Suppose these structures converge to some particular activation vector
    $\alpha \in \mathbb{R}^d$
with increasing training steps, that is, $\gamma(n) \to \alpha$.
Moreover, suppose this convergence proceeds as a power law in squared error,
\begin{equation}
    \label{eq:powerlaw}
    \|\gamma(n) - \alpha\|_2^2 = A n^{-E}
\end{equation}
for some $A, E > 0$.
Let
    $\beta \in \mathbb{R}^d$
represent an encoding of the corresponding idealised Bayesian belief
distributions.
If $\beta = \alpha$ then we have that the transformer's representations
converge to encoding Bayesian belief distributions with power law squared
error.
However, we might suppose instead that
    $\beta = \alpha + \delta$
for some small distance $\delta \in \mathbb{R}^d$.
Then, how does squared representation error (measured against the ideal
$\beta$) scale?
We have by \eqref{eq:powerlaw}
\begin{equation}
    \|\beta - \gamma(n)\|^2_2
    = \|\alpha - \gamma(n) + \delta\|^2_2
    = A n^{-E} - 2 (\gamma(n) - \alpha) \cdot \delta + \|\delta\|_2^2.
\end{equation}
For $n$ small enough that $\|\delta\|_2 \ll \sqrt{An^{-E}}$, we have
    $\|\beta - \gamma(n)\|^2_2 \approx A n^{-E}$
by Cauchy--Schwarz. That is, representation error exhibits an approximate power
law early on.
As $n \to \infty$, we have a plateau,
   $\|\beta - \gamma(n)\|^2_2 \to \|\delta\|^2_2$.
In between, if $(\gamma(n)-\alpha)\cdot\delta > \tfrac12An^{-E}$, the
trajectory $\gamma(n)$ bypasses $\beta$ as it slowly progresses towards
$\alpha$, and we have a dip.

\Cref{fig:results2} (right) illustrates this dynamic. This model is simple enough
that we were able to derive it after seeing the non-monotonicity (at about
50~million steps) and qualitatively predict the approximate plateau in
\cref{fig:results2} (left) before it emerged.

\section{Future work}

More careful analysis is needed to determine if these trends in performance and
representation error are robustly described by power laws. If there are power
laws, each appears to have multiple regimes with different exponents. All we
claim for now is that scaling is smooth, continually improving, and
\emph{roughly} power-law-shaped, for the first $\approx$30~million training
steps.

Moreover, we establish only a correlation between representation error and
prediction error scaling. It is not yet clear whether the smooth improvements
in representations \emph{cause} the smooth improvement in performance. Future
work should develop methodologies for studying the connection between these
observations.

The toy model at the end of \cref{sec:experiments} is merely illustrative. If
we make the simplifying assumption of a constant angle subtended by $\delta$
and $\gamma(n)-\alpha$ and fit the remaining parameters, the model fails to
quantitatively match the observed trends. This suggests that the
representations may not be approaching $\alpha$ as a strict power law, though
the nascent plateau suggests some kind of convergence.

Finally, it remains to explore trends in representations with increasing
parameters (number of layers, number of embedding dimensions), such as those
studied by \citet{Kaplan+2020} for behavioural scaling laws. Additional
parameters may facilitate more refined internal representations.

\section{Related work}

A related example of principled variation in behaviour in deep learning comes
from the literature on behavioural ``phase transitions'' in neural networks
as a function of training time or data composition.
Examples can be found across deep reinforcement learning
    \citetext{e.g.,
        \citealp{Cobbe+2019},
        \citealp{Cobbe+2020},
        \citealp{Langosco+2022},
        \citealp{AbdelSadek+2025},
        \citealp{Elliott+2026}%
    },
synthetic sequence modelling
    \citetext{e.g.,
        \citealp{Chen+2024},
        \citealp{Singh+2023},
        \citealp{Raventos+2023},
        \citealp{Wang+2024icl1},
        \citealp{Panwar+2024},
        \citealp{Carroll+2025},
        \citealp{Edelman+2024},
        \citealp{Park+2025}%
    },
and language modelling
    \citetext{e.g.,
        \citealp{Qin+2024},
        \citealp{Hoogland+2025icl1},
        \citealp{Wang+2025differentiation}%
    }.
Moreover, there are examples of interpretable internal representations and
mechanisms arising in \emph{structural} phase transitions, including induction
heads
    \citetext{%
        \citealp{Elhage+2021},
        \citealp{Olsson+2022},
        \citealp{Wang+2024icl1},
        \citealp{Edelman+2024}%
    },
toy models of superposition
    \citetext{%
        \citealp{Elhage+2022},
        \citealp{Chen+2023tms}%
    },
and grokking
    \citetext{%
        \citealp{Power+2022},
        \citealp{Nanda+Lieberum2022},
        \citealp{Nanda+2023grokking}%
    }.
A promising approach to reconciling the principles connecting training dynamics
and data composition is developmental interpretability, particularly via
singular learning theory
    \citetext{see, e.g.,
        \citealp{Chen+2023tms},
        \citealp{Carroll+2025},
        \citealp{Wang+2025differentiation},
        \citealp{PepinLehalleur+2025}%
    }.
So far, the implications of \emph{scale} for the formation of internal
structure have been under-explored.

\section{Conclusion}

The two most striking empirical phenomena of the modern deep learning era are
scaling laws and examples of emergent computational structure. Neither of these
phenomena is yet clearly understood, undermining experts' ability to predict
the near-future trajectory and implications of AI technology, and society's
ability to steer that trajectory.

We have proposed that these two phenomena are both reflections of some deeper
principles driving variations in the emergence and refinement of computational
structure in response to variations in scale. If so, understanding scaling laws
and emergent mechanisms alike will require first understanding the principles
relating structure and scale.

In this direction, we have taken a small first step by studying the variation
in a known internal representation mechanism while scaling compute and data. In
this simplified setting, we identified an example of a correlation between
behavioural improvements with scale and the measurable and non-monotonic
refinement of the linear encoding of Bayesian belief distributions over latent
states.
This is a hint that there is more to behavioural scaling patterns than tabular
function approximation.


\clearpage
\section*{Acknowledgements}

We thank 
    Thomas Bush,
    Pranav Mahajan,
    Daniel Murfet,
    Louis Thomson,
    and Joan Velja
for helpful conversations.
Claude Opus 4.6 and 4.7 assisted with experimentation and plot generation.
TPUs provided by Google's TPU Research Cloud.

\bibliography{main}

\clearpage
\appendix

\section{Plots without smoothing}
\label{apx:smoothing}

\begin{figure}[!b]
    \centering
    \includegraphics{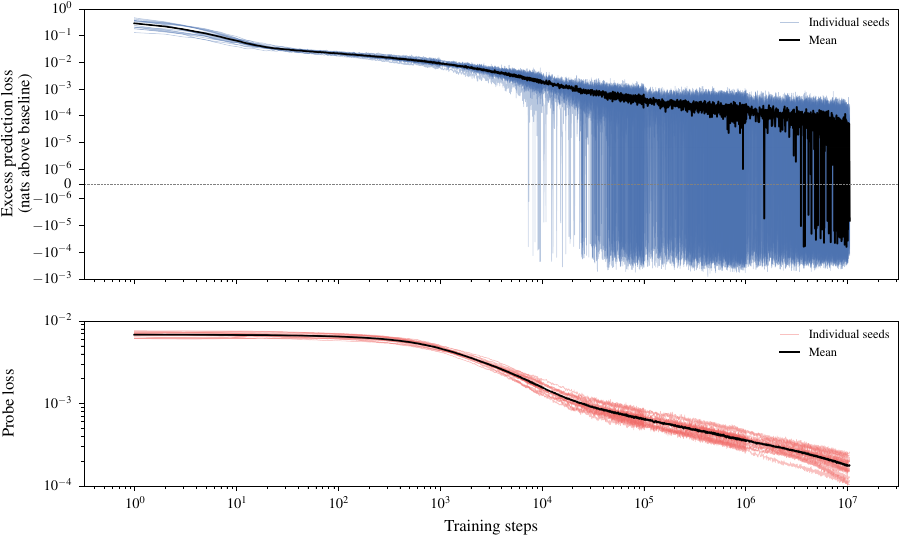}
    \caption{%
        \textbf{Raw performance/representation scaling of simplicial sequence
        models.}
        \textit{Top:} Per-step held-out test cross-entropy minus per-seed
        irreducible cross-entropy. Mean (black) of 16 seeds (blue), no
        smoothing. Vertical axis uses a symmetric log scale to display negative
        entries below the singularity.
        \textit{Bottom:} Mean squared error predicting Bayesian belief
        distributions from final-layer activations. Mean (black) of 16 seeds
        (red), no smoothing.
    }
    \label{fig:raw-results1}
\end{figure}

\Cref{fig:results1} incorporates Gaussian smoothing in log-step space
with a kernel of width $\sigma = 1/20$ decades.
\Cref{fig:raw-results1} shows the unsmoothed equivalent.

The buildup of samples towards the right side of each log-space decade reflects
our measurement schedule. We capture observables for each of the first 1,000
steps, and thereafter at approximately 1,000 linearly-spaced steps per decade.

Moreover, this plot emphasises that the excess prediction loss occasionally
drops below zero, saturating the log scale. We use a symmetric log scale to
show that (1) the loss descends only modestly below the baseline and (2) it
eventually returns positive.

Negative excess prediction loss indicates that the transformers are making
better predictions than the Bayesian posterior predictive distribution on the
fixed batch of 1024 evaluation sequences. We note that though this is
impossible \emph{in expectation over the data generating process,} it is
entirely possible for the transformer to out-predict Bayes on any given
sequence.

\end{document}